\documentclass[11pt,draftcls,onecolumn,peerreview]{IEEEtran}
\ifCLASSINFOpdf
\else
\fi

\usepackage{epsfig}
\usepackage{epstopdf}
\usepackage{amssymb}
\usepackage{amsmath}

\begin{document}
%
\title{Large Scale Distributed Acoustic Modeling With Back-off N-grams}
%
%
%

\author{Ciprian Chelba* and Peng Xu and Fernando Pereira and Thomas Richardson
\thanks{Copyright \copyright 2013 IEEE. Personal use of this material is permitted. However, permission to use this material for any other purposes must be obtained from the IEEE by sending a request to pubs-permissions@ieee.org.}
\thanks{Ciprian Chelba and Peng Xu and Fernando Pereira are with
  Google, Inc., 1600 Amphiteatre Pkwy, Mountain View, CA 94043, USA.}%
\thanks{Thomas Richardson is with the Statistics Department, Box
  354322, University of Washington, Seattle, WA 98195, USA.}
\thanks{Corresponding author: Ciprian Chelba, ciprianchelba@google.com.}}

%
%

\markboth{Journal of IEEE Transactions on Audio, Speech, and Language Processing}%
{Shell \MakeLowercase{\textit{et al.}}: Bare Demo of IEEEtran.cls for Journals}
%



\maketitle

\begin{abstract}
  The paper revives an older approach to acoustic modeling that borrows
  from n-gram language modeling in an attempt to scale up both the
  amount of training data and model size (as measured by the number of
  parameters in the model), to approximately 100 times larger than
  current sizes used in automatic speech recognition.
  In such a data-rich setting, we can expand the phonetic context
  significantly beyond triphones, as well as increase the number of
  Gaussian mixture components for the context-dependent states that
  allow it. We have experimented with contexts that span seven or more
  context-independent phones, and up to 620 mixture components per state.
  Dealing with unseen phonetic contexts is accomplished using the
  familiar back-off technique used in language modeling due to
  implementation simplicity. The back-off acoustic model is
  estimated, stored and served using MapReduce distributed
  computing infrastructure.

  Speech recognition experiments are carried out in an N-best list
  rescoring framework for Google Voice Search.
  Training big models on large amounts of data proves to be an
  effective way to increase the accuracy of a state-of-the-art automatic speech recognition
  system. We use 87\,000 hours of training data (speech along with
  transcription) obtained by filtering utterances in Voice Search logs
  on automatic speech recognition confidence. Models ranging in size between 20--40 million
  Gaussians are estimated using maximum likelihood training. They
  achieve relative reductions in word-error-rate of 11\% and 6\%
  when combined with first-pass models trained using maximum likelihood, and boosted
  maximum mutual information, respectively. Increasing the
  context size beyond five phones (quinphones) does not help.
\end{abstract}


%
\IEEEpeerreviewmaketitle

\section{Introduction}
As web-centric computing has grown over the last decade, there has
been an explosion in the amount of data available for training
acoustic and language models for speech recognition. Machine
translation~\cite{brants-EtAl:2007:EMNLP-CoNLL2007} and language
modeling for Google Voice Search~\cite{chelba:slt2010} have shown that using
more training data is quite beneficial for improving the performance
of statistical language models. The same holds true in many other applications as
highlighted in~\cite{halevy2009unreasonable}. Of equal importance is
the observation that the increase in training data amount should
be paired with an increase in the model size. This is the situation
in language modeling, where word n-grams are the core features
of the model and more training data leads to more parameters in the
model. We propose a similar approach for automatic speech recognition (ASR) 
acoustic modeling that is conceptually simpler than established techniques, 
but more aggressive in this respect.

As a first step, it is worth asking how many training samples are needed to
estimate a Gaussian well? Appendix~\ref{sec:appendix}
provides an answer for unidimensional data under the assumption that
the $n$ i.i.d. samples are drawn from a normal distribution of unknown
mean and variance, $\mathcal{N}(\mu,\sigma^2)$. We can place an upper-bound on the
probability that the sample mean $\overline{X} = \frac{1}{n}
\sum_{i=1}^n X_i$ is more than $q \cdot \sigma$ away from the actual
mean $\mu$, for $q$ small.
For example, $P(|\overline{X}-\mu| > 0.06 \cdot \sigma) < 0.06$ when
$n=983$; similar values are obtained for the sample variance estimate.

Typical amounts of training data used for the acoustic model (AM) in
ASR vary from 100 to 1000 hours. The frame rate in most systems is
100 Hz, (corresponding to advancing the analysis window in
10-millisecond steps), which means that about 360 million samples are used to
train the 0.5 million or-so Gaussians in a common state-of-the-art ASR
system. Assuming that $n=1000$ frames are sufficient for robustly
estimating a single Gaussian, then 1000 hours of speech would allow
for training about 0.36 million Gaussians. This figure is quite close
to values encountered in ASR practice, see
Section~\ref{sec:first_pass_am}, or Table VI
in~\cite{gales_et_al_BN_progress_TASLP_2006}. We can thus say that
current AMs achieve \emph{estimation efficiency}: the training data is
fully utilized for robust estimation of model parameters.

Recent applications have led to availability of data far beyond that
commonly used in ASR systems. Filtering utterances logged by the 
Google Voice Search service at an adequate ASR confidence threshold,
(see~\cite{jiang:confidence} for an overview on various confidence
measures for ASR), guarantees transcriptions that are close to human
annotator performance, e.g. we can obtain 87\,000 hours of
automatically transcribed speech at a confidence level of 0.8 or higher
in the accuracy of the transcription. If we are to strive for
estimation efficiency, then this much speech data would allow
training of AMs whose size is about 40 million Gaussians. From a
modeling point of view the question becomes: \emph{what is the best
  way to ``invest'' these parameters to meaningfully model the speech
  signal?}

The most common technique for dealing with data sparsity when
estimating context-dependent output distributions for HMM states is the well-known 
decision-tree (DT) clustering approach~\cite{young94:_treebased_state_tying}. To
make sure the clustered states have enough data for reliable
estimation, the algorithm guarantees a minimum number of frames at each
context-dependent state (leaf of the DT). The data at each
leaf is modeled by a Gaussian mixture model (GMM). At the other end of
the spectrum, states for which there is a lot more training data
should have more mixture components. There is a vast amount
of literature on such model selection techniques,
see~\cite{watanabe_et_al} for a recent approach, as well as an
overview. \cite{kim2003recent} shows that an effective way
of sizing the GMM output distribution in HMMs as a function of the
amount of training data (number of frames $n$) is the log-linear rule:
\begin{eqnarray}
  \log(\mathrm{num.\ components}) & = & \log(\beta) + \alpha \cdot \log(n)\label{eq:no_mix}
\end{eqnarray}

We take the view that we should estimate as many Gaussian components as the
data allows for a given state, according to the robustness
considerations in Appendix~\ref{sec:appendix}. In
practice one enforces both lower and upper thresholds on the number of
frames for a given GMM (see Section~\ref{sec:ml_exps} for actual
values used in our experiments), and thus the parameters $\alpha$ and
$\beta$ in (\ref{eq:no_mix}) are set such that the output
distributions for states are estimated reliably across the full range
of the data availability spectrum.

As a first direction towards increasing the model size when using
larger amounts of training data, we choose to use longer phonetic context
than the traditional triphones or quinphones: the phonetic context for an HMM
state is determined by $M$ context-independent (CI) phones to the left and
right of the current phone and state. We experiment with values for
$M = 1, \ldots, 3$, thus reaching the equivalent of $7$-phones.
For such large values of $M$ not all \emph{M-phones} (context
dependent HMM states in our model), are encountered in the training
data. At test time we deal with such unseen M-phones by backing-off,
similar to what is done in n-gram language modeling: the context for
an unseen M-phone encountered on test data is decreased gradually
until we reach an M-phone that we have already observed in training.

The next section describes our approach to increasing the state space
using back-off acoustic modeling, and contrasts it with prior work. 
Section~\ref{sec:distributed_am} describes the back-off acoustic model (BAM)
implementation using Google's distributed infrastructure, primarily
MapReduce~\cite{Dean:2008:MSD:1327452.1327492} and SSTable (immutable
persistent B-tree, see \cite{chang2008bigtable}), along similar lines
to their use in large scale language modeling for statistical machine
translation~\cite{brants-EtAl:2007:EMNLP-CoNLL2007}.
Section~\ref{sec:experiments} presents our experiments in an N-best list
rescoring framework, followed by conclusions. The
current paper is a more comprehensive description of the experiments reported
in~\cite{chelba:icassp2012_am}.

\section{Back-off N-grams for Acoustic Modeling}
\label{sec:bam}
Consider a short utterance whose transcription is: 
$\boldsymbol{W} =\ $\verb+<S> action </S>+, and assume the pronunciation lexicon
provides the following mapping to CI phones
\verb+sil ae k sh ih n sil+. We use \verb+<S>, </S>+ to denote sentence
boundaries, both pronounced as long silence \verb+sil+.

A triphone approach would model the three states of \verb+ih+ as
\verb$sh-ih+n_{1,2,3}$ using the DT clustering algorithm
for tying parameters across various instances
\verb$*-ih+*_{1,2,3}$, respectively. This yields the so-called
context-dependent states in the HMM.

In contrast, BAM with $M=3$ extracts the following training data
instances (including back-off) for the first HMM state of the
\verb+ih+ instance in the example utterance above:\\
\verb+ih_1 / ae k sh ___ n sil    frames+\\
\verb+ih_1 /    k sh ___ n sil    frames+\\
\verb+ih_1 /      sh ___ n        frames+\\
There are other possible back-off strategies, but we currently
implement only the one above:
\begin{itemize}
\item if the M-phone is symmetric (same left and right context length),
then back-off at both ends
\item if not, then back-off from the longer end until the M-phone becomes
  symmetric, and proceed with symmetric back-offs from there on.
\end{itemize}
To achieve this we first compute the context-dependent state-level Viterbi
alignment between transcription $\boldsymbol{W}$ and speech feature frames using the
transducer composition $\boldsymbol{H} \circ  \boldsymbol{C} \circ
\boldsymbol{L} \circ \boldsymbol{W}$, where $\boldsymbol{L},\ \boldsymbol{C},\
\boldsymbol{H}$ denote respectively the pronunciation lexicon, context dependency
tree, and HMM-to-state FST transducers~\cite{mohri2002weighted}. From the
alignment we then extract M-phones along
with the corresponding sequence of speech feature \verb+frames+. 
Each M-phone is uniquely identified by its \emph{key},
e.g. \verb+ih_1 / ae k sh ___ n sil+. The key is a string
representation obtained by joining on \verb+ / + the central CI-state,
i.e. \verb+ih_1+ above, and the surrounding phonetic context, in this
case \verb+ae k sh ___ n sil+; \verb+___+ is a placeholder marking the
position where the central CI-state \verb+ih_1+ occurs in the context.
Besides the maximal order M-phones, we also collect \emph{back-off}
M-phones as outlined above. With each back-off we clone the frames
from the maximal order M-phone to the back-off one. We found it useful
to augment the phonetic context with word boundary information. The
word boundary has its own symbol, and occupies its own context
position.

All M-phone instances encountered in the training data are aggregated using
MapReduce. For each M-phone that meets a
threshold on the minimum number of frames aligned against itself, we estimate a GMM
using the standard splitting algorithm~\cite{htk_book}, following the
rule in (\ref{eq:no_mix}) to size the GMM.  The M-phones that
have more frames than an upper threshold on the maximum number frames
(256k in our experiments),\footnote{For convenience, we use the ``k''
  shorthand to denote thousands, e.g. we write 256k instead of
  256\,000; a value of 41\,898\,799 is rounded to 41\,899k.} are
estimated using reservoir sampling~\cite{vitter1985random}. Variances
that become too low are floored to a small value (0.00001 in our
experiments).

\subsection{Comparison with Existing Approaches and Scalability Considerations}
BAM can be viewed as a simplified version of DT state clustering that
uses question sets consisting of atomic CI phones,
queried in a pre-defined order. This very likely makes BAM sub-optimal
relative to standard DT modeling, yet we prefer it due to ease of
implementation in MapReduce.

The approach is not novel:~\cite{schwartz1984improved} proposes a very
similar strategy where the probability assigned to a frame by a
triphone GMM is interpolated with probabilities assigned by left,
right diphone GMMs, and CI phone GMMs, respectively.
However, the modeling approach in BAM is not identical
to~\cite{schwartz1984improved} either: the former does indeed back-off
in that it uses only the maximum order M-phone found in the model,
whereas the latter interpolates up the back-off tree, and allows
asymmetric back-offs. It is of course perfectly
feasible to conceive BAM variants that come closer to the approach
in~\cite{schwartz1984improved} by using interpolation between M-phones
at various orders.

Scalability reasons make the current BAM implementation an easier
first attempt when using very large amounts of training data: a BAM
with $M=5$ estimated on 87\,000 hours of training data leads to
roughly 2.5 billion (2\,489\,054\,034) 11-phone types. 
DT building requires as sufficient statistics the single-mixture
Gaussians for M-phones sharing the same central
CI phone and state. Assuming uniformity across central
phone and state identity, we divide the total number of M-phones by
the number of phones (40) times the number of 
states/phone (3) to arrive at about 25 million different
M-phones that share a given central state and phone. Storing a
single-mixture Gaussian for each M-phone requires
approximately $320$ bytes ($39\cdot4\cdot2$). Under the uniformity assumption above,
the training data for each DT amounts to about $25\cdot320 = 8$~GB
of storage. It is more realistic
to assume that some central CI phones will have ten times more
M-phones than the average, leading to a memory footprint of 80
GB, which starts to become problematic although still feasible (perhaps
by employing sampling techniques, or reducing the context size $M$).

To avoid such scalability issues, we resort to MapReduce and streaming
the data for M-phones sharing a given central triphone to the same
Reducer, one maximal order M-phone at a time, as described in
Section~\ref{sec:distributed_am}. M-phones at lower orders $1 \ldots
M-1$ are estimated by accumulating the data arriving at the Reducer into buffers of
fixed capacity, using reservoir sampling~\cite{vitter1985random} to
guarantee a random sample of fixed size of the input data stream. 
With careful sorting of the M-phone stream (see Section~\ref{sec:distributed_am}),
$M-1$ reservoirs are sufficient for buffering data on the Reducer
until all the data for a given M-phone
arrives, the final GMM for a given M-phone is estimated and output,
and the respective buffer is flushed. The reservoir size thus controls
the memory usage very effectively. For example, when using 256k as the
maximum number of frames for estimating a given GMM (equal to the
maximum reservoir size), only 160~MB of RAM are sufficient for
building a BAM with $M=5$.

Our approach to obtaining large amounts of training data is very
similar to that adopted in~\cite{gales_et_al_BN_progress_TASLP_2006}. Table VI there
highlights the gains from using increasing amounts of training data
from 375 hours to 2210 hours, and shows that past 1350 hours a
system with 9k states and about 300k Gaussians gets diminishing
returns in accuracy. Our modeling approach and its implementation
using MapReduce allows both the use of significantly more
training data and estimation of much larger models: in our experiments
we used 87\,000 hours of training data and built models of up to 1.1
million states and 40 million Gaussians.

\section{Distributed Acoustic Modeling}
\label{sec:distributed_am}
BAM estimation and run-time are implemented using MapReduce and
SSTable, and draw heavily from the large language modeling approach
for statistical machine translation described
in~\cite{brants-EtAl:2007:EMNLP-CoNLL2007}.

\subsection{BAM Estimation Using MapReduce}
\label{bam_mr}
Our implementation is guided by the large scale n-gram language model estimation work of~\cite{brants:distributed}.
MapReduce is a framework for parallel processing across huge datasets
using a large number of machines. The computation is split in
two phases: a \emph{Map} phase, and a \emph{Reduce} one.
The input data is assumed to be a large collection of key-value
pairs residing on disk, and stored in a distributed file
system. MapReduce divides it up into \emph{chunks}. Each such chunk
is processed by a Map worker called a \emph{Mapper}, running on
a single machine, and whose entire lifetime is dedicated to processing
one such data chunk.

Mappers are stateless, and for each input key-value pair in a given
chunk they output one or more new key-value pairs; the
computation of the new value, and the assignment of the new output
key are left to the user code implementing a Mapper instance. 
The entire key space of the Map output is disjointly
partitioned according to a \emph{sharding function}: for any key value
output by Map, we can identify exactly one Reduce \emph{shard}. 

The key-value pairs output by all Mapper instances are routed to their
corresponding Reduce shards by a \emph{Shuffler}, using the
sharding function mentioned above. The Shuffler also collates all
values for a given key, and presents the tuple of values along with
the key to the \emph{Reducer} (a Reduce worker), as one of the many
inputs for a given Reduce shard. It also sorts the keys for a given
shard in lexicographic order, which is the order in which they are
presented to the Reducer. Each Reduce worker processes all the key-value pairs for a given
Reduce shard: the Reducer receives a key along with all the values
associated with it as output by all the Mappers, and collected by the
Shuffler; for a given input key, the Reducer processes all values
associated with it, and outputs one single key-value pair. It is worth
noting that the Reducer cannot change the identity of the key it
receives as input. The output from the Reduce phase is stored in an
\emph{SSTable}: an immutable persistent B-tree\footnote{A format
  similar to SSTable has been open-sourced as part of the LevelDB
  project http://code.google.com/p/leveldb/} associative array of
key-value pairs (both key and value are strings), that is partitioned
according to the same sharding function as the one used by the MapReduce
that produced it. Another SSTable feature is that it can be used
as a distributed in-memory key-value serving system (\emph{SSTable service}) with
$S$ servers (machines), each holding a partition containing $1/S$ of
the total amount of data. This allows us to serve models larger
than what would fit into the memory of a single
machine. 

\begin{figure*}[!t]
  \centerline{\includegraphics[width=0.95\linewidth]{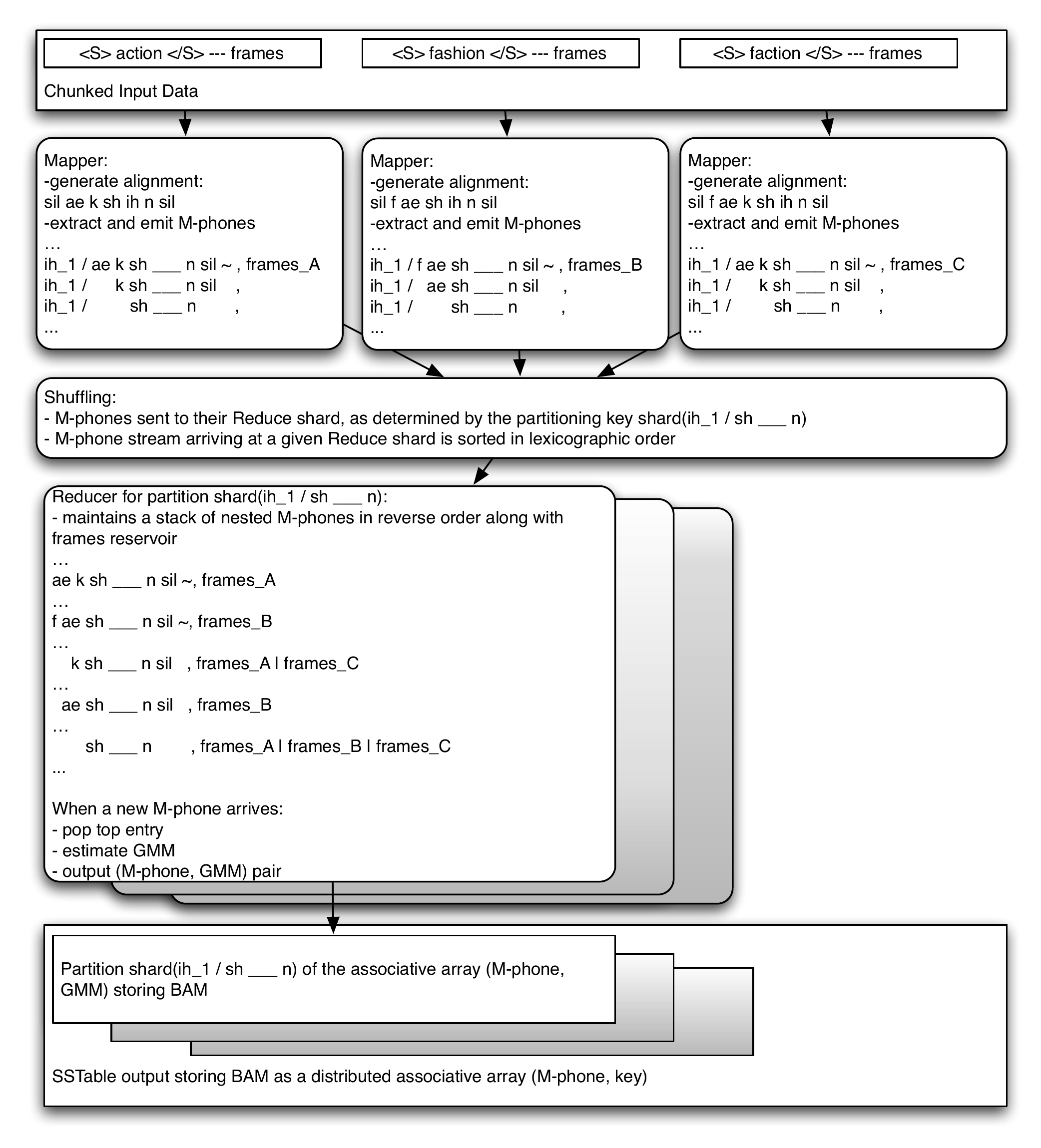}}
  \caption{MapReduce Estimation for Back-Off Acoustic Model}
  \label{fig:mr}
\end{figure*}

Fig.~\ref{fig:mr} describes the training MapReduce, explained
in more detail in the following two subsections.

\subsubsection{Mapper}
\label{mapper}
Each Mapper instance processes a chunk of the input data, one record
at a time. Each record consists of a key-value pair; the value stores
the waveform, the word level transcript for the utterance, and other
elements. For each record arriving at the Mapper we:
\begin{itemize}
\item generate the context-dependent state-level Viterbi alignment by
  finding the least cost path through the state space of the FST 
  $\boldsymbol{H} \circ \boldsymbol{C} \circ \boldsymbol{L} \circ
  \boldsymbol{W}$ using the the first-pass AM
\item extract maximal order M-phones along with speech frames, and
  output (M-phone key, frames) pairs
\item compute back-off M-phones and output (M-phone key, empty) pairs.
\end{itemize}
We note that in order to avoid sending too much data to the Reducer,
we do not copy the frames to the back-off M-phones, which would lead
to replicating the input data M times. To make sure that the data
needed for estimating back-off M-phones is present at a given Reducer
we resort to a few tricks:
\begin{itemize}
\item the sharding function takes as argument the central triphone. This
  guarantees that all M-phones sharing a given central triphone
  (\verb$sh-ih+n$ in our example), are handled by the same Reducer
\item the M-phones need to arrive at the Reducer in a certain order,
  since only the maximal order M-phones carry speech frame data. The
  sorting of the M-phone stream needs to be such that any given
  maximal order M-phone arrives at the Reducer before all of its
  back-off M-phones; this allows us to buffer the
  frames for all the back-off M-phones down to the central triphone
  state. We accomplish this by relying on the implicit lexicographic sorting of the keys,
  and re-keying each M-phone before outputting it such that the
  context CI-phones are listed in proximity order to the central
  one; missing context CI-phones (due to utterance boundaries), are
  represented using \verb+~+ to ensure correct sorting. For example
  \verb+ih_1 / ae k sh ___ n sil+ is actually keyed as: 
  \verb+ih_1 / sh n k sil ae ~+, to guarantee that M-phones sharing 
  the central triphone \verb+ih_1 / sh ___ n+ are processed in order 
  of longest to shortest context at the Reducer processing 
  the partition \verb+partition(ih_1 / sh ___ n)+.
\end{itemize}

\subsubsection{Reducer}

After shuffling, each M-phone has its frame data (if carrying any),
collated and presented to the Reducer along with the M-phone
key. Since the Reducer cannot change the key of the input, it needs to
output the GMM for an M-phone when it arrives at the Reducer. The
sorting described in Section~\ref{mapper} guarantees that the
M-phones sharing the same central triphone arrive in the correct
order (high to low M-phone order). Every time a maximal order
M-phone arrives at the Reducer we estimate a GMM from its data
(assuming the number of frames is above the lower threshold), and also
accumulate its data in the \emph{reservoirs} for all of its back-off
M-phones which are buffered in a ``first-in first-out'' stack.

Reservoir sampling is a family of randomized algorithms for randomly choosing
$K$ samples from a list $L$ containing $n$ items, where $n$ is either
a very large or unknown number. Our implementation populates the
reservoir with the first $K$ samples to arrive at the
Reducer. If more samples arrive after that, we draw a random
index $r$ in the range 
$[0, \mathrm{current\ sample\ index} - 1)$; if $r < K$ we replace the
sample at index $r$ in the reservoir with the newly arrived one, and
otherwise we ignore it.
Every time a back-off M-phone arrives at the Reducer, it
is guaranteed to be the same one as the M-phone at the top of the
stack due to the sorting of the M-phone stream done by the
Shuffler. We then:
\begin{itemize}
\item add to the reservoir at the top of the stack any frames that
  arrived at the Reducer with the current M-phone;
\item pop the M-phone and the corresponding reservoir from the top of the
stack;
\item estimate the GMM for this back-off M-phone if the
accumulated frames exceed the lower threshold on the minimum number of
frames, or discard the M-phone and its data otherwise;
\item output the pair (M-phone key, GMM).
\end{itemize}

Due to the particular sorting of the M-phone stream, the Reducer is
guaranteed to have seen all the frame data for an M-phone when the
GMM estimation takes place. The resulting SSTable stores the
BAM as a distributed (partitioned) associative array (M-phone key, GMM).

\subsection{BAM Test Run-time Using SSTable Service}

At test time we rescore N-best lists for each utterance using BAM. We
load the model into an SSTable service with $S$ servers, each
holding $1/S$ of the data. For each hypothesis in the N-best list, we:
\begin{itemize}
\item generate the context-dependent state-level Viterbi alignment after composing
  $\boldsymbol{H} \circ \boldsymbol{C} \circ \boldsymbol{L}$ with the transcript $\boldsymbol{W}$ from the first-pass; the alignment is
  generated using the first-pass AM and saved with the hypothesis
\item extract maximal order M-phones
\item compute back-off M-phones
\item add all M-phones to a pool initialized once per input record (utterance).
\end{itemize}

Once the pool is finalized, it is sent as a batch request to the
SSTable service. The M-phones that are actually stored in the model
are returned to the Mapper, and are used to rescore the alignment for
each of the hypotheses in the N-best list. For each segment in the
alignment we use the highest order M-phone that was retrieved from
the BAM SSTable. If no back-off M-phones are retrieved for a given
segment, we back-off to the first-pass AM score for that
segment which is computed during the Viterbi alignment.

To penalize the use of lower order M-phones, 
the score for a segment with an M-phone of lower order $o$ ($o \geq 0$) than the
maximum one $M$ incurs a per-frame back-off cost. The order of an asymmetric M-phone is computed as the
maximum of the left and right context lengths. The per-frame back-off
cost reaches its maximum value when the model
backs-off all the way to using the first-pass AM (DT clustered state), 
$o = 0$. To formalize, assume that we are using a GMM with $Q$ components for
modeling M-phone $s$, and that the order of $s$ is $o(s)$,
computed as described above. The log-likelihood assigned to a frame
$\boldsymbol{y}$ aligned against state $s$ will be:
\begin{eqnarray}
\log P_\mathrm{s}(\boldsymbol{y}) 
& = & \log \sum_{q=1}^Q m_q \cdot P(\boldsymbol{y}|\boldsymbol{\mu}_{s,q},\boldsymbol{\Sigma}_{s,q}) - \nonumber\\
&   & f_\mathrm{bo} \cdot (M - o(s))\label{eq:fbo}
\end{eqnarray}
where $f_\mathrm{bo} \geq 0$ is the per-frame back-off cost, and $m_q$
are the mixture weights for each component of the GMM for state $s$:
$P(\boldsymbol{y}|\boldsymbol{\mu}_{s,q},\boldsymbol{\Sigma}_{s,q})$.

The final score for each hypothesis $\boldsymbol{W}$, $\log P(\boldsymbol{W},\boldsymbol{A},\boldsymbol{V})$, is computed by
log-linear interpolation between the first-pass AM and that obtained
from the second pass one (BAM, or first-pass AM if running sanity checks, see
Table~\ref{tab:ml_results}), followed by the usual log-linear combination
between AM and language model (LM) scores:
\begin{eqnarray}
\log P_\mathrm{AM}(\boldsymbol{A}|\boldsymbol{W},\boldsymbol{V}) & = & \lambda \cdot \log P_\mathrm{1st\ pass}(\boldsymbol{A}|\boldsymbol{W},\boldsymbol{V}) + \nonumber\\
                 &   & (1.0 - \lambda) \cdot \log P_\mathrm{2nd\
                   pass}(\boldsymbol{A}|\boldsymbol{W},\boldsymbol{V}) - \nonumber\\
                 &   & \log(Z_1) \label{eq:log-lin-am}\\
\log P(\boldsymbol{W},\boldsymbol{A},\boldsymbol{V})    & = & 1/w_{LM} \cdot \log P_\mathrm{AM}(\boldsymbol{A}|\boldsymbol{W},\boldsymbol{V}) + \nonumber\\
                 &   & \log P_\mathrm{LM}(\boldsymbol{W}) - \log(Z_2), \label{eq:log-lin}
\end{eqnarray}
where $\boldsymbol{A}$ denotes the acoustic features, $\boldsymbol{W}$ denotes the word
sequence in an N-best hypothesis, $w_{LM}$ is the language model weight, $P(\boldsymbol{W},\boldsymbol{A},\boldsymbol{V})$ is the
probability assigned to the word sequence $\boldsymbol{W}$ and the corresponding
acoustic features $\boldsymbol{A}$ by using $\boldsymbol{V}$ as a state-level alignment, and
$\log(Z_1)$, $\log(Z_2)$ are normalization terms ignored in rescoring;
both first-pass AM and BAM pair states with frames using the same 
state-level Viterbi alignment $\boldsymbol{V}$ computed using the first-pass AM.

\section{Experiments}
\label{sec:experiments}
We ran our experiments on Google Voice Search training and test
data. The subsections below detail the training and test setup, as well as
the baseline acoustic models and their performance.

\subsection{Task Description}
There are two training sets that we used in our experiments:
\begin{itemize}
\item \emph{maximum likelihood (ML) baseline}: 1 million manually transcribed Voice Search spoken
  queries, consisting of 1300 hours of speech (468\,887\,097 frames);
\item \emph{filtered logs}: 110 million Voice Search spoken queries
  along with 1-best ASR transcript,
  filtered by confidence at 0.8 threshold, consisting of 87\,000 hours of
  speech (31\,530\,373\,291 frames). The query-level confidence score used for filtering training
  data transcriptions is derived using standard lattice-based word
  posteriors. The best baseline AM available, namely the
  boosted maximum mutual information (bMMI) baseline AM trained as we describe in
  Section~\ref{sec:first_pass_am} is used for generating 
  both transcriptions and confidence scores.
\end{itemize}

As development and test data we used two sets of manually transcribed data
that do not overlap with the training data (the utterances originate
from non-overlapping time periods in our logs). Let's denote them as
data sets DEV, and TEST, consisting of 27\,273 and 26\,722 spoken queries
(87\,360 and 84\,918 words), respectively. All query data
used in the experiments (training, development and test), is anonymized.

As a first attempt at evaluating BAM, we carry out N-best list rescoring
experiments with $N=10$. While 10-best may seem small, such N-best lists
have approximately 7\% oracle word-error-rate (WER)\footnote{The oracle WER measures the
WER along the hypothesis in the N-best list that is closest in string-edit distance to the
transcription for that utterance.} on our development data set,
starting from 15\% WER baseline. Also, as shown in
Section~\ref{exp:validation}, about 80\% of the test set achieves 0\%
oracle WER at 10-best, so there is plenty of room for improvement when
doing 10-best list rescoring. In addition to this, very large LM rescoring
experiments for the same task, e.g.~\cite{chelba:slt2010}, have shown
that 10-best list rescoring was very close to full lattice rescoring.

\subsection{First Pass Acoustic Models}
\label{sec:first_pass_am}
The feature extraction front-end is common across all experiments:
\begin{itemize}
\item the speech signal is sampled at 8 kHz, and quantized linearly on 16 bits
\item 13-dimensional perceptual linear predictive (PLP)
  coefficients~\cite{hermansky:plp} are extracted every 10 ms using a
  raised cosine analysis window of size 25 ms, consisting of 200 samples
  zero-padded to 512 samples
\item 9 consecutive PLP frames around the current one are
  then stacked to form a 117-dimensional vector
\item a joint transformation estimated using linear discriminant
  analysis (LDA) followed by semi-tied covariance (STC)
  modeling~\cite{Gales99semi-tiedcovariance} reduces the feature
  vector down to 39 dimensions in a way that minimizes the loss from
  modeling the data with a diagonal covariance Gaussian distribution.
\end{itemize}

Since BAM uses ML estimation, we decided to use two
baseline AMs in our experiments: an ML baseline AM that
matches BAM training, and a discriminative (bMMI) baseline AM
which produces the best available results on our development and test
data. All models use diagonal covariance Gaussians.

The ML AM used in the first-pass is estimated on the ML baseline data
in the usual staged approach:
\begin{enumerate}
\item three-state, CI phone HMMs with output
  distributions consisting of single Gaussian, diagonal
  covariance
\item standard DT clustering for triphones, producing 8k context-dependent states
\item GMM splitting, resulting in a model with 330k Gaussians:
  \begin{itemize}
  \item the minimum number of frames $N_\mathrm{min}$ for a given
    context-dependent state is 18k, enforced during DT building;
  \item the maximum number of frames $N_\mathrm{max}$ for a given
    context-dependent state is 256k; GMMs for states with more than
    the maximum number of frames are estimated by random sampling down
    to 256k frames
  \item \emph{varmix} estimation is used to determine the number of
    mixtures according to the amount of training data, as in (\ref{eq:no_mix})
    with $\alpha=0.3, \beta=2.2$; this amounts to 42 components when the number of
    frames $n$ is at its minimum value of 18k, and 92 mixture
    components when it is at its maximum value of 256k.
  \end{itemize}
\end{enumerate}

The bMMI baseline AM is obtained by running an additional discriminative
training stage on significantly more training data than the ML baseline:
\begin{enumerate}
\setcounter{enumi}{3}
\item bMMI training~\cite{bMMI} on the ML baseline data
  augmented with 10 million Voice Search spoken queries (approximately
  8000 hours) and 1-best ASR transcript, filtered by confidence.
\end{enumerate}

Training and test are matched with respect to the first-pass AM used:
experiments reporting development and test data results using the ML
baseline AM use a BAM trained on alignments generated using the same ML
baseline AM; likewise, when switching to the bMMI baseline AM we use it
to generate training, development and test data alignments.

\subsection{N-best List Rescoring Experiments using ML Baseline AM}
\label{sec:ml_exps}

The development data is used to optimize the following parameters for
BAMs trained on the ML baseline data, as well as 1\%, 10\% and
100\% of the filtered logs data, respectively:
\begin{itemize}
\item model order $M = 1, 2, \ldots, 3$ (triphones to 7-phones),
\item acoustic model weight in log-linear mixing of first-pass AM scores
  with the rescoring AM, (\ref{eq:log-lin-am}): $\lambda = 0.0, 0.2, 0.4, \ldots, 1.0$,
\item language model weight, (\ref{eq:log-lin}): $w_{LM} = 7, 12, \ldots, 22$,
\item per-frame back-off weight, (\ref{eq:fbo}): $f_\mathrm{bo} = 0.0,
  0.2, \ldots, 1.0$.
\end{itemize}

Across all experiments reported in this section we kept the following constant:
\begin{itemize}
\item the ML baseline AM is trained on the ML baseline data,
\item minimum number of frames for an M-phone state
  $N_\mathrm{min}$ is 4k except for one experimental condition setting
  it to 18k to compare against the ML baseline AM, see Table~\ref{tab:ml_results},
\item maximum number of frames (reservoir size), $N_\mathrm{max}$ for
  an M-phone state is 256k:
  \begin{itemize}
  \item for the $\alpha = 0.3$ and $\beta = 2.2$ varmix setting this means a maximum
    number of 92 mixture components per state
  \item for the $\alpha = 0.7$ and $\beta = 0.1$ varmix setting this means 620
    mixture components per state; the GMM splitting becomes very slow
    for such large numbers of mixture components, so we only trained $M = 1$
    models for this setting.
  \end{itemize}
\end{itemize}

\subsubsection{Development Set Results}
\begin{table*}[!t]
  \centering

  \caption{Maximum Likelihood Back-off Acoustic Model (BAM) Results on
    the Development Set, 10-best Rescoring, in Various Training and
    Test Regimes.}
  \label{tab:ml_results_dev}

  \begin{small}
    \begin{tabular}{|l|r|r|}\hline
      Model                                                & WER (S/D/I), & No. \\
                                                           &         (\%) & Gaussians\\\hline
      \multicolumn{3} {|l|} {\underline{\emph{TRAINING DATA = ML baseline data (1.3k hours)}}} \\
      ML baseline AM, $\lambda=0.0$, $w_{LM}=17$              & 19.1 (2.3/4.3/12.5) & 327k\\
      ML baseline AM, $\lambda=0.6$, $w_{LM}=17$              & \emph{18.5} (2.2/4.2/12.0) & 327k\\
      ML baseline AM, $\lambda=1.0$ (first-pass), $w_{LM}=17$  & \underline{18.8} (2.3/4.3/12.2) & 327k\\\hline
      \multicolumn{3} {|l|} {\underline{\emph{TRAINING DATA = 100\% filtered logs data (87k hours)}}} \\
      BAM $N_\mathrm{min}=4k$, $\alpha=0.3$, $\beta=2.2$, $M=1$,
      $\lambda=0.0$, $w_{LM}=17$, $f_\mathrm{bo} = 0.0$ & 18.0 (2.4/3.9/11.7) & 3213k\\
      BAM $N_\mathrm{min}=4k$, $\alpha=0.3$, $\beta=2.2$, $M=1$,
      $\lambda=0.6$, $w_{LM}=17$, $f_\mathrm{bo} = 0.0$ & 17.1 (2.2/3.8/11.1) & 3213k\\\hline
      BAM $N_\mathrm{min}=4k$, $\alpha=0.3$, $\beta=2.2$, $M=2$,
      $\lambda=0.0$, $w_{LM}=17$, $f_\mathrm{bo} = 0.0$ & 17.7 (2.0/4.2/11.6) & 22\,210k\\
      BAM $N_\mathrm{min}=4k$, $\alpha=0.3$, $\beta=2.2$, $M=2$,
      $\lambda=0.6$, $w_{LM}=17$, $f_\mathrm{bo} = 0.0$ & 16.8 (2.0/3.9/10.9) & 22\,210k\\\hline
      BAM $N_\mathrm{min}=4k$, $\alpha=0.3$, $\beta=2.2$, $M=3$,
      $\lambda=0.0$, $w_{LM}=17$, $f_\mathrm{bo} = 0.0$ & 18.0 (2.0/4.2/11.8) & 41\,899k\\
      BAM $N_\mathrm{min}=4k$, $\alpha=0.3$, $\beta=2.2$, $M=3$,
      $\lambda=0.6$, $w_{LM}=17$, $f_\mathrm{bo} = 0.0$ & 16.9 (2.0/3.9/11.0) & 41\,899k\\
      BAM $N_\mathrm{min}=4k$, $\alpha=0.3$, $\beta=2.2$, $M=3$,
      $\lambda=0.6$, $w_{LM}=17$, $f_\mathrm{bo} = 0.2$ & 16.8 (2.0/3.8/11.0) & 41\,899k\\
      BAM $N_\mathrm{min}=4k$, $\alpha=0.3$, $\beta=2.2$, $M=3$,
      $\lambda=0.6$, $w_{LM}=17$, $f_\mathrm{bo} = 0.4$ & 16.8 (2.0/3.8/11.0) & 41\,899k\\
      BAM $N_\mathrm{min}=4k$, $\alpha=0.3$, $\beta=2.2$, $M=3$,
      $\lambda=0.6$, $w_{LM}=17$, $f_\mathrm{bo} = 0.6$ & 16.8 (2.0/3.8/11.0) & 41\,899k\\
      BAM $N_\mathrm{min}=4k$, $\alpha=0.3$, $\beta=2.2$, $M=3$,
      $\lambda=0.6$, $w_{LM}=17$, $f_\mathrm{bo} = 0.8$ & 16.9 (2.0/3.8/11.1) & 41\,899k\\
      BAM $N_\mathrm{min}=4k$, $\alpha=0.3$, $\beta=2.2$, $M=3$,
      $\lambda=0.6$, $w_{LM}=17$, $f_\mathrm{bo} = 1.0$ & 16.9 (2.0/3.8/11.1) & 41\,899k\\\hline
    \end{tabular}
  \end{small}
\end{table*}

Table~\ref{tab:ml_results_dev} shows the most relevant results when
rescoring 10-best lists with BAM in the log-linear interpolation
 (\ref{eq:log-lin}); S/D/I denotes Substitutions, Deletions
and Insertions, respectively. 

We built and evaluated models for  $M = 1, 2, \ldots, 5$ but as the results in
Table~\ref{tab:ml_results_dev} show, there is no gain in performance
for values of $M > 2$; since training such models is expensive, we
stopped early on experimenting with models at $M=4, 5$, and as such
Table~\ref{tab:ml_results_dev} reports results for $M = 1, 2, 3$ only.

The first three rows show the performance, and size (in number
of Gaussians), of the ML AM baseline (stage 3 in
Section~\ref{sec:first_pass_am}) on the development set
DEV. Somewhat surprisingly, there is a small gain (0.3\% absolute)
obtained by interpolating the first and second pass scores produced by
the ML baseline AM for the same utterance, as well as a loss of 0.3\%
absolute when the N-best list is rescored with the same AM. We point out this
oddity because the same second pass alignments are rescored with
BAM, and hence this small improvement should not be credited to better
modeling using BAM, but rather to re-computation of alignments in
the second pass for each N-best hypothesis individually.
This discrepancy could be due to one or more possible sources of mismatch
between the first-pass system and the N-best list rescoring one\footnote{We
  tried our best to minimize this discrepancy but given the many
  parameter settings in an ASR system this task has proven to be very
  difficult. The small difference reported was the best we could achieve
  after spending a significant amount of time on this issue.}:
\begin{itemize}
\item different frame level alignments for the same word
  hypothesis. This could happen due to the fact that the rescoring
  system uses extremely wide beams when computing the alignment for
  each hypothesis in the N-best list, as well as the fact that some
  optimizations in the generation of the first-pass static $\boldsymbol{CLG}$ FST
  network may not be matched when aligning a given hypothesis
  $\boldsymbol{W}$ using $\boldsymbol{H} \circ \boldsymbol{C} \circ \boldsymbol{L} \circ
  \boldsymbol{W}$;
\item slightly different acoustic model settings in computing the
  log-likelihood of a frame;
\item slightly different front-end configurations between first-pass
  and rescoring.
\end{itemize}

The per-frame back-off (\ref{eq:fbo}) does not make any difference at
all for $M=1, 2$ models (we do not include the results in
Table~\ref{tab:ml_results_dev} since they are identical to those
obtained under the $f_\mathrm{bo} = 0.0$ condition), and has a minimal
impact on the $M=3$ model.

Another point worth making is that BAM stands on its own, at least in
the N-best list rescoring framework investigated:
comparing the rows for $\lambda=0.0$, we observe that BAM improves
over the ML baseline AM for all values $M=1, 2, 3$, with the optimal
value being $M=2$.

Finally, the fact that the larger $M=3$ value does not improve
performance over the $M=2$ model despite the availability of data to
estimate GMMs reliably is an interesting result in its own
right, suggesting that simply increasing the phonetic context beyond
quinphones may in fact weaken the acoustic model.

\subsubsection{Test Set Results}

\begin{table*}[!t]
  \centering

  \caption{Maximum Likelihood Back-off Acoustic Model (BAM) Results on
    the Test Set TEST, 10-best list Rescoring, in Various Training
    Regimes.}
  \label{tab:ml_results}

  \begin{small}
    \begin{tabular}{|l|r|r|}\hline
      Model & WER (S/D/I), & No. \\
            & (\%)         & Gaussians\\\hline
      \multicolumn{3} {|l|} {\underline{\emph{TRAINING DATA = ML baseline data (1.3k hours)}}} \\
      ML baseline AM, $\lambda=0.0$, $w_{LM}=17$              & 12.4 (1.3/2.5/8.6) & 327k\\
      ML baseline AM, $\lambda=0.6$, $w_{LM}=17$              & \emph{11.6} (1.2/2.3/8.1) & 327k\\
      ML baseline AM, $\lambda=1.0$ (first-pass), $w_{LM}=17$  & \underline{11.9} (1.2/2.4/8.3) & 327k\\\hline
      \multicolumn{3} {|l|} {\underline{\emph{TRAINING DATA = ML baseline data (1.3k hours)}}} \\
      BAM $N_\mathrm{min}=18k$, $\alpha=0.3$, $\beta=2.2$, $M=1$,
      $\lambda=0.8$, $w_{LM}=17$, $f_\mathrm{bo} = 0.0$ & 11.6 (1.2/2.2/8.2) & 223k\\
      BAM $N_\mathrm{min}=4k$, $\alpha=0.3$, $\beta=2.2$, $M=1$,
      $\lambda=0.8$, $w_{LM}=17$, $f_\mathrm{bo} = 0.0$ & 11.5 (1.2/2.2/8.1) & 490k\\\hline
      \multicolumn{3} {|l|} {\underline{\emph{TRAINING DATA = 1\% filtered logs data (870 hours)}}} \\
      BAM $N_\mathrm{min}=4k$, $\alpha=0.3$, $\beta=2.2$, $M=2$,
      $\lambda=0.8$, $w_{LM}=17$, $f_\mathrm{bo} = 1.0$ & 11.3 (1.2/2.2/7.9) & 600k\\
      BAM $N_\mathrm{min}=4k$, $\alpha=0.7$, $\beta=0.1$, $M=1$,
      $\lambda=0.8$, $w_{LM}=12$, $f_\mathrm{bo} = 0.0$ & 11.4 (1.1/2.3/8.0) & 720k\\\hline

      \multicolumn{3} {|l|} {\underline{\emph{TRAINING DATA = 10\% filtered logs data (8.7k hours)}}} \\
      BAM $N_\mathrm{min}=4k$, $\alpha=0.3$, $\beta=2.2$, $M=2$,
       $\lambda=0.6$, $w_{LM}=17$, $f_\mathrm{bo} = 0.4$ & 10.9 (1.1/2.2/7.7) & 3975k\\
      BAM $N_\mathrm{min}=4k$, $\alpha=0.7$, $\beta=0.1$, $M=1$,
      $\lambda=0.6$, $w_{LM}=17$, $f_\mathrm{bo} = 0.0$ & 10.9 (1.1/2.2/7.6) & 4465k\\\hline

      \multicolumn{3} {|l|} {\underline{\emph{TRAINING DATA = 100\% filtered logs data (87k hours)}}} \\
      BAM $N_\mathrm{min}=4k$, $\alpha=0.3$, $\beta=2.2$, $M=2$,
      $\lambda=0.6$, $w_{LM}=17$, $f_\mathrm{bo} = 0.0$ & \bf{10.6} (1.0/2.2/7.4) & 22\,210k\\
      BAM $N_\mathrm{min}=4k$, $\alpha=0.7$, $\beta=0.1$, $M=1$,
      $\lambda=0.6$, $w_{LM}=17$, $f_\mathrm{bo} = 0.0$ & \bf{10.6} (1.2/2.0/7.3) & 14\,435k\\\hline
    \end{tabular}
  \end{small}
\end{table*}

Table~\ref{tab:ml_results} shows the results of rescoring 10-best lists
with the BAM in the log-linear interpolation setup of
(\ref{eq:log-lin}), along with the best settings as estimated on the
development data.

The first training regime for BAM used the same training data as that
used for the ML part of the baseline AM training sequence. When
matching the threshold on the minimum number of frames to the threshold used for the
baseline AM (18k), BAM ends up with fewer Gaussians than the baseline
AM: 223k vs. 327k. This is not surprising, since no DT clustering is
done, and the data is not used as effectively: many triphones (i.e., $M=1$) are
discarded, along with their data. However, its performance matches that
of the baseline AM in a 10-best list rescoring setup; no claims
are made about the performance of such a model in the first
pass. Lowering the threshold on the minimum number of frames to 4k (26 mixture
components at $\alpha=0.3$, $\beta=2.2$), does increase the number of
Gaussians in the model to 490k.

The second training regime for BAM uses the filtered logs data,
in varying amounts: 1\%, 10\%, 100\%, respectively. A surprising
result is that switching from manually annotated data to the same
amount of confidence filtered data provides a small absolute WER gain
of 0.1--0.2\%. This suggests that the confidence filtered data is just
as good as the manually annotated data for training acoustic models
that are used in an N-best list rescoring pass.

\begin{figure*}[!t]
  \centerline{\includegraphics[width=0.8\linewidth]{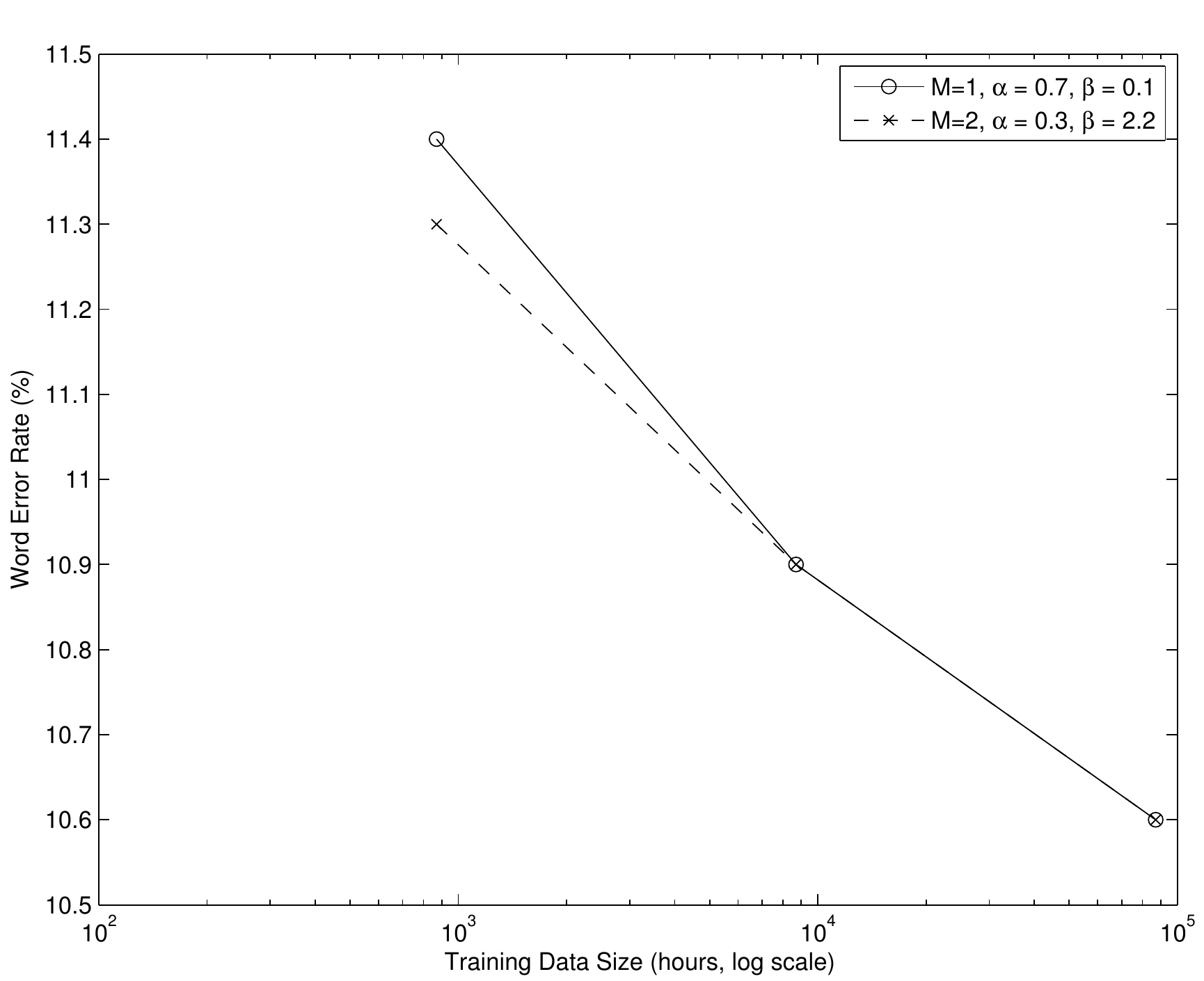}}
  \caption{ASR Word Error Rate as a Function of Training Data Size, for two different BAM configurations. The WER decreases almost linearly with the logarithm of the training data size, and it is marginally influenced by the BAM order (context size). }
  \label{fig:wer_vs_data}
\end{figure*}

From then on, BAM steadily improves as we add more filtered logs training data in both the
$\alpha=0.3$, $\beta=2.2$ and $\alpha=0.7$, $\beta=0.1$ setups, respectively: the first
ten-fold increase in training data brings a 0.4--0.5\% absolute WER
reduction, and the second one brings another 0.3\% absolute WER
reduction. As shown in Fig.~\ref{fig:wer_vs_data}, the WER decreases
almost linearly with the logarithm of the training data size. 

The BAM WER gain amounts to 1.3\% absolute reduction (11\% relative) on the one-pass baseline of
11.9\% WER. Comparing the baseline results when using ML and bMMI
models, respectively (see
Tables~\ref{tab:ml_results}~and~\ref{tab:mmi_results}), we note that BAM does
not fully close the 18\% relative difference between the ML and the
bMMI first-pass AMs performance, which leaves open the possibility that a
discriminatively trained BAM would yield additional accuracy gains.
\begin{figure*}[!t]
  \centerline{\includegraphics[width=0.8\linewidth]{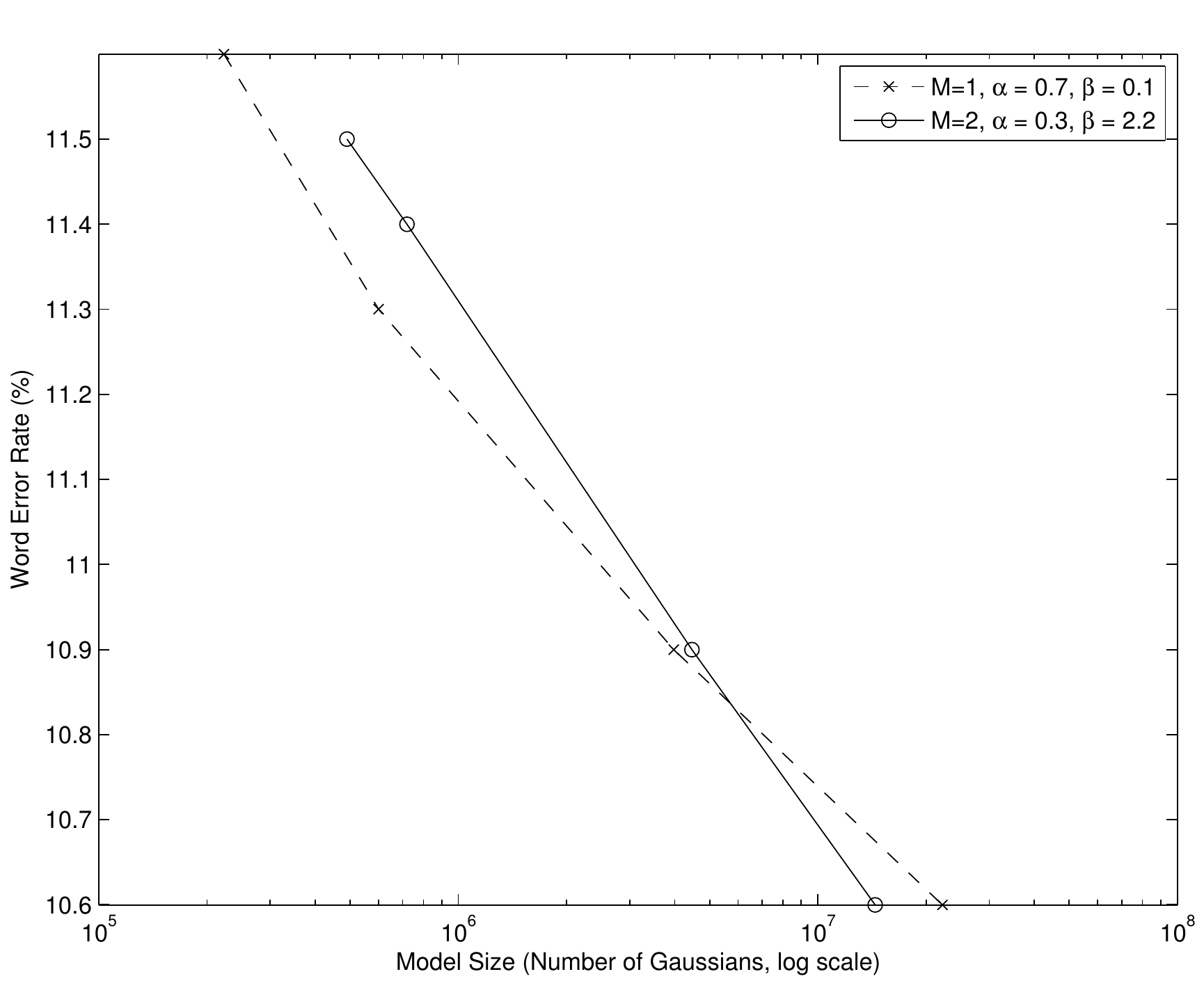}}
  \caption{ASR Word Error Rate as a Function of Model Size, for two different BAM configurations. The WER decreases almost linearly with the logarithm of the model size (measured in number of Gaussians), and it is marginally influenced by the BAM order (context size).}
  \label{fig:wer_vs_model}
\end{figure*}

As Fig.~\ref{fig:wer_vs_model} shows, the best predictor for model
performance is the number of mixture components, which is
consistent with the results on development data, and across the two
different $\alpha, \beta$ settings we experimented with. The best
model order $M$ is between $M=1$ and $M=3$
(depending on the maximum number of mixtures/state allowed in the model).
In fact, with enough mixtures per M-phone, triphones (i.e., $M=1$) perform just
as well as quinphones (i.e., $M=2$) or 7-phones.

\subsection{N-best List Rescoring Experiments using bMMI Baseline AM}
\label{sec:mmi_exps}
When switching to using the bMMI AM (stage 4 in
Section~\ref{sec:first_pass_am}) as the first-pass model in
both training and test, the baseline results are significantly better,
see Table~\ref{tab:mmi_results}. Despite the fact that it is not discriminatively
trained, BAM still provides 0.6\% absolute (6\% relative) reduction in WER.
\begin{table*}[!t]
  \centering

  \caption{Discriminative (boosted-MMI) Acoustic Model Baseline Results and
    BAM Performance on the Test Set TEST, 10-best List Rescoring.}
  \label{tab:mmi_results}

  \begin{small}
    \begin{tabular}{|l|r|r|r|r|}\hline
      Model                                                & WER (S/D/I), & No. \\
                                                           &         (\%) & Gaussians\\\hline
      \multicolumn{3} {|l|} {\underline{\emph{TRAINING DATA = ML baseline data (1.3k hours) + 10k hours filtered logs data}}}\\
      bMMI baseline AM, $\lambda=0.0$, $w_{LM}=17$             & 10.2 (1.1/1.7/7.4) & 327k\\
      bMMI baseline AM, $\lambda=0.6$, $w_{LM}=17$             &  \emph{9.7} (1.1/1.6/7.0) & 327k\\
      bMMI baseline AM, $\lambda=1.0$ (first-pass), $w_{LM}=17$ &  \underline{9.8} (1.1/1.6/7.1) & 327k\\\hline
      \multicolumn{3} {|l|} {\underline{\emph{TRAINING DATA = 100\% filtered logs data (87k hours)}}} \\
      BAM $N_\mathrm{min}=4k$, $\alpha=0.3$, $\beta=2.2$, $M=3$,
      $\lambda=0.8$, $w_{LM}=17$, $f_\mathrm{bo} = 0.0$) & \bf{9.2} (1.0/1.6/6.7) & 40\,360k\\\hline
    \end{tabular}
  \end{small}
\end{table*}

\subsection{M-phone Hit Ratios and Other  Training Statistics}
Similar to n-gram language modeling, we can compute M-phone \emph{hit
  ratios} at various orders: the percentage of M-phones
encountered in the test data (10-best hypotheses), with left, right
context of length $l, r$, respectively, that are
also present in the model (and thus there is no need to back-off further);
Table~\ref{tab:bo_hit_ratios} shows the values for BAM trained on the
filtered logs data (87\,000 hours). M-phones at query boundaries do not
have symmetric context, which explains the non-zero off-diagonal values.
The maximal order M-phones (sum across last row and column), amount to
42.3\% of the total number of M-phones encountered on 10-best list
rescoring, with 23.6\% at the highest order 3, 3. 

We also note that only on 1.1\% of test segments do we back-off out of the M-phones
stored in BAM, and use the GMM stored with the clustered state in
the first-pass AM. This shows convincingly that as both the amount of
training data and the model size increase, the DT
clustering of triphone states is no longer necessary as a means to
cope with triphones that are unseen or have too little training data.
\begin{table}[!h]
\centering

\caption{M-phone Hit Ratios on 10-best Hypotheses for Test Data for
  BAM Using $M=3$ (7-phones) Trained on the Filtered Logs Data (87\,000 hours)}

\label{tab:bo_hit_ratios}

\begin{small}
  \begin{tabular}{|l|r|r|r|r|}\hline
    left, right              &         &           &            & \\
    context size            & 0       &         1 &          2 & 3 \\\hline
    0                       & 1.1\% &  0.1\% &  0.2\% &  4.3\%\\
    1                       & 0.1\% & 26.0\% &  0.9\% &  3.4\%\\
    2                       & 0.7\% &  0.9\% & 27.7\% &  2.2\%\\
    3                       & 3.8\% &  2.9\% &  2.0\% & 23.6\%\\\hline
  \end{tabular}
\end{small}
\end{table}
Tables~\ref{tab:no_gaussians} and \ref{tab:no_m_phone_types} show the
distribution of Gaussian mixtures, and M-phone types at various
orders, respectively. 
The total number of Gaussian mixtures in the model is 41\,898\,799, and the
total number of M-phone types is 1\,146\,359, achieving
our goal of scaling the AM 100 times larger than the size of the first-pass AM, 
which consists of 0.3 million Gaussians and about 8k context-dependent states.
\begin{table}[!h]
  \centering
  
  \caption{Number of Gaussian Mixtures at Various M-phone Orders for BAM Using
    $M=3$ (7-phones) Trained on the Filtered Logs Data}
  
  \label{tab:no_gaussians}
  
  \begin{small}
    \begin{tabular}{|l|r|r|r|r|}\hline
      left, right              &        &        &         &        \\
      context                 &        &        &         &        \\
      size                    & 0      &      1 &       2 &       3\\\hline
      0                       & 0	&     9802 &    138\,694 &    776\,488\\
      1                       & 9\,619   & 3\,193\,495 &    581\,846 &  1\,072\,242\\
      2                       & 143\,940 &   613\,401 & 17\,519\,632 &  1\,134\,640\\
      3                       & 843\,282 & 1\,274\,683 &  1\,127\,789 & 13\,459\,246\\\hline
    \end{tabular}
  \end{small}
\end{table}
\begin{table}[!h]
  \centering
  
  \caption{Number of M-phone Types at Various Orders for BAM Using
    $M=3$ (7-phones) Trained on the Filtered Logs Data}
  
  \label{tab:no_m_phone_types}
  
  \begin{small}
    \begin{tabular}{|l|r|r|r|r|}\hline
      left, right             &         &         &          &         \\
      context                 &         &         &          &         \\
      size                    & 0       &       1 &        2 &        3\\\hline
      0                       & 0       &     114 &     1902 &  14\,384\\
      1                       & 115     & 55\,551 &  11\,942 &  27\,426\\
      2                       & 2124    & 12\,673 & 491\,528 &  32\,248\\
      3                       & 15\,858 & 33\,035 &  32\,717 & 414\,742\\\hline
    \end{tabular}
  \end{small}
\end{table}

\subsection{Data Flow in Training MapReduce}
\label{sec:data_flow}
The filtered logs training data consists of approximately 110 million
Voice Search spoken queries, or 87\,000 hours of speech, or 31.5
billion frames; on disk it is stored as compressed SSTables at around
5.7~TB. The output of the Map
phase consists of about 8.54~TB uncompressed data, which is processed
by the Reduce function. 

Of the total of 184 million M-phones encountered in the training
data (including back-offs), only one million pass the lower threshold on
the number of frames (4k); of those, approximately 36k (3.6\%) have more
frames than the upper threshold (256k), and are estimated using
reservoir sampling.

Most time is spent during GMM splitting in the Reduce
phase. The estimation takes about 48 hours on 1000 Reduce workers; at
half-time, there are approximately 10\% Reduce partitions still being
worked on: since we need to use our own partitioning function, the
Reduce partitions are fairly uneven, with the largest partition being about
70~GB (a lot of the data sent to the largest 3--5 reduce partitions is
silence frames), and the smallest about 2~GB.

The size on disk for the largest models we built is about 30~GB. For
N-best list rescoring we load the generated data into an in-memory
key-value serving system with 100 servers, each holding 1/100 of the
model stored uncompressed for faster look-up.

\subsection{Validation Setup}
\label{exp:validation}
To verify the correctness of our implementation we set up a
validation training and test bed:
\begin{itemize}
\item train on the development set by keeping all the data and
  M-phones by setting the minimum number of frames
  $N_\mathrm{min} = 1$;
\item test on the subset of the development data with 0\% WER
  at 10-best (the manual transcription is among the
  10-best hypotheses extracted in the first-pass). This is a significant
  part of the development set, approximately 80\% (21\,751/27\,273)
  utterances. The first-pass AM achieves 7.6\% WER on this subset;
\item minimize the effect of the language model and first-pass AM scores by setting
  $w_{LM} = 0.1, \lambda= 0.0$ in
  (\ref{eq:log-lin-am}-\ref{eq:log-lin}), and use only the AM score
  assigned by BAM.
\end{itemize}

The intent behind choosing this setup is that as the order $M$
increases, BAM should ``memorize'' the alignments on the training set
(even M-phones with a single training frame are retained in the
model), and severely penalize mismatched alignments from N-best
competitors to the correct transcript at test time. It is for this
reason that we choose to test on the subset which contains the correct
transcription (used in training) in the N-best list.

The results are presented in Table~\ref{tab:validation_results} for
various context types, and model orders $M$. A surprising result is
the fact that a triphone equivalent BAM ($M=1$) that does not use
word boundary information is significantly weaker than its counterpart
that uses that information. Increasing the model order improves
performance in both context settings. The residual WER is due to
homophones.
\begin{table}[!h]
\centering

\caption{Word-Error-Rates in Validation Setup, Using Various Context
  Types as Well as Model Orders $M$}

\label{tab:validation_results}

\begin{small}
  \begin{tabular}{|l|r|r|}\hline
    Context type            & M    & WER, (\%) \\\hline
    CI phones               & 1    & 4.5 \\
    CI phones               & 5    & 1.5 \\\hline
    \ \ \ + word boundary   & 1    & 1.8 \\
    \ \ \ + word boundary   & 5    & 0.6 \\\hline
  \end{tabular}
\end{small} 
\end{table}

\section{Conclusion and Future Work}
\label{sec:conclusion}

We find these results very encouraging, proving that large scale distributed
acoustic modeling has the potential to greatly improve the quality of ASR.
Expanding phonetic context is not really productive: ``more model'' by
increasing $M > 2$ yields no gain in accuracy, so we still need to
find alternative ways to fully exploit the large amounts of data that
are now available. The best predictor for ASR performance is the model
size, as measured by the number of Gaussians in the model.

State clustering using DTs as a means of coping with data
sparsity may no longer be necessary: only 1.1\% of the state-level
segments on test data alignments back-off to the DT
clustered GMM. It remains to be seen if DTs have other modeling
advantages: since we used a rescoring framework, and the first-pass
alignments are generated with an AM that uses DT state clustering,
it is still very much part of the core modeling approach presented. In addition
to that, our best results are obtained by interpolating BAM with the
baseline AM.

Obvious future work items that are perfectly feasible at this scale
include: DT state tying, re-computing alignments in BAM ML
training, and discriminative GMM training. 

Another possible direction exploits the BAM ability to deal with large 
phonetic contexts, and large amounts of data. In the early stages of this work 
we successfully built BAMs with $M=5$, but their performance on development data 
did not justify experimenting further with such large models. It would be 
interesting to build BAMs by starting from the surface form of words (using 
letters as context elements instead of phones) and inferring the HMM topology for 
each unit in a purely data-driven manner, along the lines described in 
Section 3.6 of~\cite{Jelinek:1998:SMS:280484}.

It seems that we literally have more data than we know what to do with, and 
better modeling techniques at large scale are needed. Non-parametric modeling
techniques may be well suited to taking advantage of such large
amounts of data.

\appendices
\section{How Much Data is Needed to Estimate a Gaussian  Well?}
\label{sec:appendix}

Consider $n$ i.i.d. samples $X_1,...,X_n$ drawn from a normal
distribution $\mathcal{N}(\mu,\sigma^2)$. We would like an upper-bound on the
probability that the sample mean estimate $\overline{X} = \frac{1}{n}
\sum_{i=1}^n X_i$ is more than $q \cdot \sigma$ away from the actual
mean, with $q \in (0,1)$.

If $X_1,...,X_n \sim \mathcal{N}(\mu,\sigma^2)$ then $(\overline{X} -
\mu)/(\sigma / \sqrt{n}) \sim \mathcal{N}(0,1)$. Thus $P(|\overline{X} - \mu| >
q \cdot \sigma) = P(|Z| > q \cdot \sqrt{n}) = 2 \cdot \Phi(-\sqrt{n} \cdot
q)$ where $Z$ is the standard normal i.e. $Z \sim \mathcal{N}(0,1)$, and $\Phi$
is the cumulative distribution function (CDF) for a standard normal
random variable. Thus $P(|\overline{X} - \mu| > q \cdot \sigma) < p$
is equivalent to choosing $n$ such that: $2 \cdot \Phi(-\sqrt{n} \cdot q) <
p$. In Matlab this can be easily calculated as\\
\verb+n = (icdf('Normal', (1-p/2), 0, 1)/q)^2+ and in R as\\
\verb+n = (qnorm(1-p/2)/q)^2+, see Table~\ref{tab:num-samples} for a
few sample values.
\begin{table}[!h]
  \centering

  \caption{Number of Samples $n$ Needed to Estimate the Mean of a Normal
    Distribution within $q \cdot \sigma$ of the Actual Mean, with
    Probability Lower than $p$}
  \label{tab:num-samples}

  \begin{small}
    \begin{tabular}{|r|r|r|}\hline
      $p$    & $q$    & $n$   \\\hline
      0.05   & 0.05   & 1537 \\
      0.06   & 0.06   & 983 \\
      0.07   & 0.07   & 670 \\
      0.08   & 0.08   & 479 \\
      0.10   & 0.10   & 271 \\
      0.15   & 0.15   & 95 \\\hline
    \end{tabular}
  \end{small}
\end{table}

A ``good'' value for the sample size is $n = 300, \ldots, 1000$.  We also
note that if the sample size is this large then the statement will
still hold approximately true even if the population is not
normal, since by the central limit theorem $\overline{X}$ will be
very close to normal even if the population is not.

A similar derivation can be carried out for the sample variance
estimate: assuming normality,\\ $(n-1) \cdot S^2/\sigma^2$ follows a $\chi^2$
distribution on $n-1$ d.f.

\section*{Acknowledgments}

Many thanks to our colleagues that helped with comments, suggestions,
and solving various speech infrastructure issues, in particular: A.
Gruenstein, B. Strope, D. Beeferman, E. McDermott, J. Dean,
J. Schalkwyk, M. Bacchiani, P. Nguyen, T. Brants,
V. Vanhoucke, and W. Neveitt. Special thanks go to O.
Siohan for help with prompt code reviews and detailed comments.




%
\bibliographystyle{IEEEtran}
\bibliography{/home/ciprianchelba/documents/latex/mainbibfile}

%

\begin{IEEEbiography}[{\includegraphics[width=1in,height=1.25in,clip,keepaspectratio]{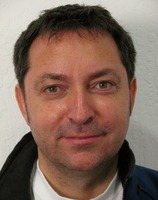}}]{Ciprian Chelba}
(Senior Member, IEEE) received his Diploma Engineer degree 
in 1993 from the Faculty of Electronics and Telecommunications at "Politehnica"
University, Bucuresti, Romania, M.S in 1996 and Ph.D. in 2000 from the
Electrical and Computer Engineering Department at the Johns Hopkins
University.

Between 2000 and 2006 he worked as a Researcher in the Speech Technology Group 
at Microsoft Research, after which he joined Google, where he is currently 
Staff Research Scientist.

His research interests are in statistical modeling of natural
language and speech, as well as related areas such as machine
learning with an emphasis on large-scale data-driven modeling. 

Recent projects include query stream language modeling for Google voice search, 
speech content indexing and ranking for search in spoken documents, 
discriminative language modeling for large vocabulary speech recognition, 
logs mining and large-scale acoustic modeling for large vocabulary speech recognition, 
language modeling for text input on soft-keyboards for mobile devices, 
as well as speech and text classification.

He is co-inventor on more than twenty US patents, many filed internationally as well. 
His publications besides numerous conference and journal papers include tutorials presented at 
HLT-NAACL 2006, ICASSP 2007, an article published in the 
IEEE Signal Processing Magazine Special Issue on Spoken Language Technology 2008, 
and three chapter contributions to edited books:
The Handbook of Computational Linguistics and Natural Language Processing (Wiley-Blackwell, 2010), 
Spoken Language Understanding: Systems for Extracting Semantic Information from Speech (Wiley, 2011), 
Mobile Speech and Advanced Natural Language Solutions (Springer, 2013).

He served as a member of the IEEE Signal Processing Society Speech and Language 
Technical Committee, guest editor for Computer, Speech and Language, and one of 
the chairs for Interspeech 2012 (Portland, OR, USA), area of 
Speech Recognition - Architecture, Search, Linguistic Components.
\end{IEEEbiography}

\begin{IEEEbiography}[{\includegraphics[width=1in,height=1.25in,clip,keepaspectratio]{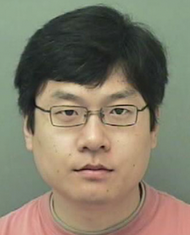}}]{Peng Xu}
received his Diploma Engineer degrees (dual-major) in 1995 from 
Tsinghua University, Beijing, China, M.S. in 1998 from Institute of Automation,
Chinese Academy of Sciences, Beijing, China, a second M.S. in 2000 and Ph.D. in 2005 from the
Electrical and Computer Engineering Department at the Johns Hopkins
University.

Since 2005, he has been working in Google Research, where he is now a senior Staff Research Scientist.
His research interests are in statistical modeling of natural
language and speech, as well as related areas such as machine
learning with an emphasis on large-scale data-driven modeling. 

Recent projects include large scale distributed language models, distributed machine translation
systems, distributed machine learning systems, dependency tree-to-string translation models,
as well as pruning techniques for language models and translation models.
\end{IEEEbiography}


\begin{IEEEbiography}[{\includegraphics[width=1in,height=1.25in,clip,keepaspectratio]{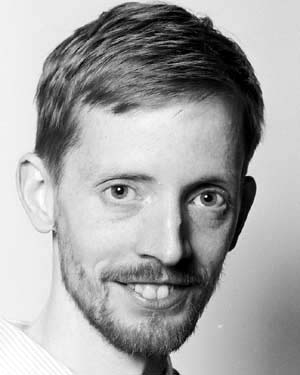}}]{Thomas Richardson}
was born in Oxford, U.K. in 1971. He received the B.A. degree (with honors) 
in mathematics and philosophy from the University of Oxford, Oxford, U.K., 
in 1992, the M.Sc. degree in logic and computation and the Ph.D. degree in logic, 
computation and methodology from Carnegie-Mellon University, Pittsburgh, PA 
in 1995, and 1996, respectively.

He joined the Department of Statistics at the University of Washington, Seattle, WA in 1996, 
where he is now a Professor. He is also the Director of the Center for Statistics and 
the Social Sciences at the University of Washington. He is the author of more than 
70 technical publications. His research interests include machine learning, 
multivariate statistics, graphical models, and causal inference. 
He has been the program co-chair for the Workshop on Artificial 
Intelligence and Statistics (AISTATs), and the Conference on Uncertainty in 
Artificial Intelligence (UAI). He has also served as an Associate Editor for the 
{\it Journal of the Royal Statistical Society Series B} and an Editor for {\it Statistical Science}.

Professor Richardson is a Fellow of the Center for Advanced Studies in the Behavioral Sciences 
at Stanford University. In 2009 he received the UAI Best Paper Award; he also received 
the Outstanding Student Paper Award at UAI in 2004 (as co-author) and in 1996 (as author).
\end{IEEEbiography}

\begin{IEEEbiography}[{\includegraphics[width=1in,height=1.25in,clip,keepaspectratio]{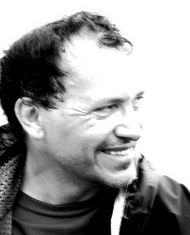}}]{Fernando Pereira} is research director at Google. His previous appointments include chair of the Computer and Information Science department at University of Pennsylvania, head of the Machine Learning and Information Retrieval department at AT\&T Labs, and research and management positions at SRI International. He received a Ph.D. in Artificial Intelligence from the University of Edinburgh in 1982. His main research interests are in machine-learnable models of language and biological sequences. He has over 100 research publications on computational linguistics, machine learning, bioinformatics, speech recognition, and logic programming, and several patents. He was elected Fellow of the American Association for Artificial Intelligence in 1991 for his contributions to computational linguistics and logic programming, and he was president of the Association for Computational Linguistics.
\end{IEEEbiography}




\end{document}